\documentclass{article}
\usepackage{spconf,amsmath,graphicx}
\usepackage{cite}
\usepackage{epstopdf}
\usepackage{xcolor}
\usepackage{amssymb}
\usepackage{pifont}
\usepackage{array}
\usepackage{booktabs}
\usepackage{bm}
\usepackage{multirow}
\usepackage{graphicx}
\usepackage{color}
\usepackage{float}
\usepackage{stfloats}
\usepackage[misc]{ifsym}
\usepackage{stmaryrd}
\usepackage{grffile}
\usepackage{subfig}
\usepackage{makecell}
\usepackage{threeparttable}
\usepackage{enumitem}
\usepackage{enumerate}
\usepackage{xspace}
\usepackage{pifont}
\usepackage{hyperref}
\DeclareUnicodeCharacter{2212}{-}

\title{Sub-Aperture Feature Adaptation in Single Image Super-resolution Model for Light Field Imaging}
\name{Aupendu Kar, Suresh Nehra, Jayanta Mukhopadhyay, Prabir Kumar Biswas}
\address{Indian Institute of Technology Kharagpur, India}
\begin{document}
%
\maketitle
\begin{abstract}

With the availability of commercial Light Field (LF) cameras, LF imaging has emerged as an up-and-coming technology in computational photography. However, the spatial resolution is significantly constrained in commercial micro-lens-based LF cameras because of the inherent multiplexing of spatial and angular information. Therefore, it becomes the main bottleneck for other applications of light field cameras. This paper proposes an adaptation module in a pre-trained Single Image Super-Resolution (SISR) network to leverage the powerful SISR model instead of using highly engineered light field imaging domain-specific Super Resolution models. The adaption module consists of a Sub-aperture Shift block and a fusion block. It is an adaptation in the SISR network to further exploit the spatial and angular information in LF images to improve the super-resolution performance. Experimental validation shows that the proposed method outperforms existing light field super-resolution algorithms. It also achieves PSNR gains of more than $1$ dB across all the datasets as compared to the same pre-trained SISR models for scale factor $2$, and PSNR gains $0.6−1$ dB for scale factor $4$.
\end{abstract}
\vspace{-5pt}
\section{Introduction}
\label{sec:intro}

A Light Field (LF) camera not only provides spatial information but also captures the angular information of the incoming light from a scene point. Therefore, it enables the LF camera to improve image processing performance in different applications such as depth estimation~\cite{peng2020zero}, image segmentation~\cite{khan2019view}, image editing~\cite{shon2016spatio}, and many more. These techniques can be further improved if we have an image of higher spatial resolution. In the case of an LF camera, multiplexing of angular and spatial information results in poor spatial resolution of Sub-Aperture (SA) images. For example, the Lytro Illum sensor resolution is $40$ MP, but a sub-aperture image's spatial resolution is close to $0.1$ MP. Therefore it is necessary to achieve Super-Resolution (SR) in LF image by exploiting the additional angular information present in the LF data.
\begin{figure}
    \centering
    \includegraphics[width=0.45\textwidth]{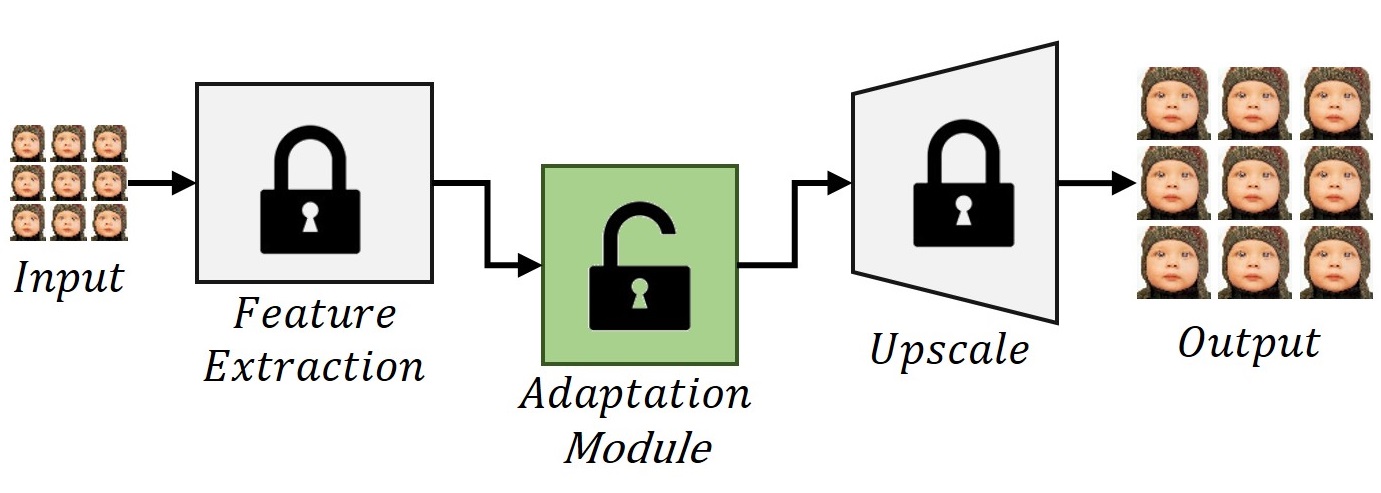}
    \vspace{-5pt}
    \caption{Feature extraction and upscale module are from any pre-trained SISR model. The adaptation module aims to learn more information from multiple sub-aperture images. During training, the weights of the adaptation module are updated (shown as an `unlocked' symbol), and the weights of two pre-trained modules are fixed (shown using `locked' symbols).}
    \label{fig: algo}
    \vspace{-15pt}
\end{figure}

Recently, Light Field Super-Resolution (LFSR) has been an area of active research, and a considerable amount of work has been done in the last few years. The earlier works were mainly based on Bayesian or variational optimization frameworks with different priors such as variational model~\cite{wanner2013variational}, Gaussian mixture model~\cite{mitra2012light}, and PCA analysis model~\cite{Farrugia2017}. These methods are inefficient in exploiting Spatio-angular information of the light field data. In contrast, learning-based methods have been proposed to achieve SR via cascaded convolutions and data-driven training. Single Image Super-Resolution (SISR) methods~\cite{VDSR, EDSR, RDN, RCAN} are becoming increasingly deep and complicated with improved capability in spatial information exploitation. However, the angular information of LF images remains unexploited in SISR networks,  resulting in limited performance. Inspired by learning-based methods in SISR and in pursuit of exploiting the angular information, recent LFSR methods~\cite{resLF, LFSSR, LF-InterNet, wang2020light} adopted deep Convolution Neural Networks (CNNs) to improve SR performance. 

This paper proposes a novel Light Field Sub-Aperture Feature Adaptation (LFSAFA) module and puts it into a pre-trained single image super-resolution model for achieving LFSR. LFSAFA exploits the angular information present in the SA images of LF data to improve the performance of LFSR. The proposed module consists of Sub Aperture Shift (SAS) and feature fusion blocks. SAS blocks process the Sub-Aperture (SA) features, and the fusion block combines those features. The modulated SA features are then fed to the upscaling network to reconstruct high-resolution images. Our experimental validation shows that pre-trained SISR models with simple LF-specific modifications can perform better than highly engineered light field image-specific super-resolution models. To summarize, the contributions of this proposed work are as follows.
\begin{itemize}[noitemsep, nolistsep]
    \item We propose a light-field domain adaptation module to achieve LFSR using SISR models. To the best of our knowledge, this is the first work in this direction. 
    \item We show that the proposed module can utilize angular information present in SA images to improve the performance, and ablation studies support our claims.
    \item Our qualitative and quantitative analysis shows that the performance of our method is better than light-field domain-specific super-resolution solutions, and any SISR models can adopt our proposed modification to make it work for LFSR.
\end{itemize}

\vspace{-5pt}
\section{Related Work}
\label{sec:prior_work}
Due to the advancement of deep learning architectures and algorithms, the LFSR domain has witnessed tremendous progress. C. Yoon et al.~\cite{LFCNN2017} super-resolved Sub-Aperture Images (SAIs) via CNN and then fine-tuned using angular information to enhance both spatial and angular resolutions. LF-DCNN~\cite{LF-DCNN} super-resolved each SAI via a more powerful SISR network and fine-tuned the initial results using an EPI-enhancement network. LFNet~\cite{LFNet} proposed a bidirectional recurrent network by extending BRCNN~\cite{BRCN}. Zhang et al.~\cite{resLF} proposed a multi-stream residual network (resLF) by stacking SAIs as inputs to super-resolve the center-view SAI. LFSSR~\cite{LFSSR} alternately shuffle LF features between SAI pattern and macro-pixel image pattern for convolution. Jin et al.~\cite{jin2020learning} proposed an all-to-one geometric aware method using structural consistency regularization that preserves the parallax structure among reconstructed views. LF-InterNet~\cite{LF-InterNet} used spatial-angular information interaction for LFSR. LF-DFnet~\cite{wang2020light} performed feature alignment using an angular deformable alignment module (ADAM). MEG-Net~\cite{zhang2021end} considered multiple epipolar geometry, and all views are simultaneously super-resolved through an end-to-end manner. DPT~\cite{DPT} used SAIs as a sequence and introduced a detail-preserving transformer to learn geometric dependencies among those sequences.

\begin{figure}[!ht]
    \centering
    \includegraphics[width=0.47\textwidth]{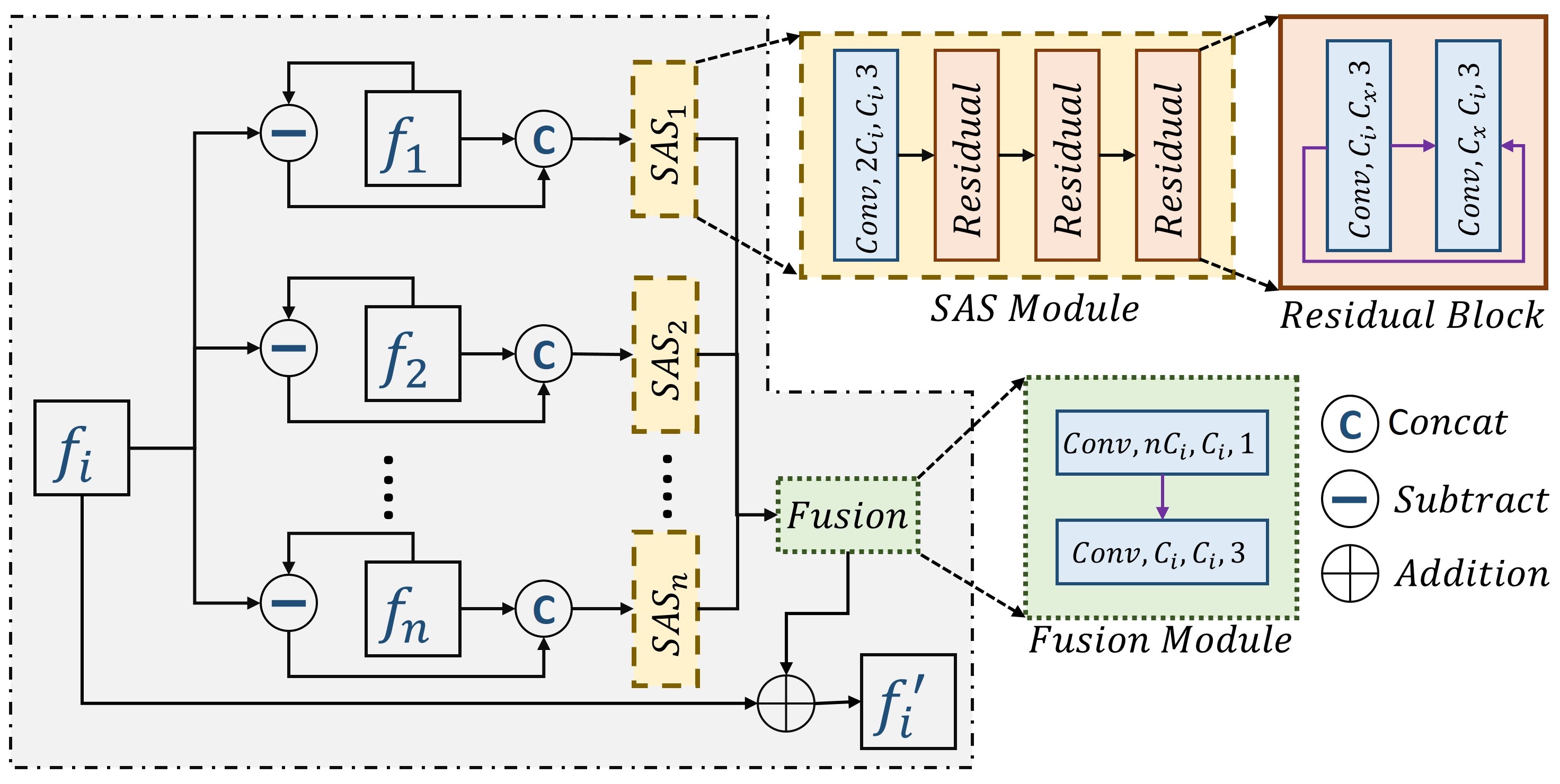}
    \caption{The proposed sub-aperture feature adaptation module consists of $n$ SAS modules and $1$ fusion module. $f_i$ is the extracted feature of a sub-aperture image using a pre-trained model $F_{feat}$ and $f_i^{'}$ is the modulated feature that contains more rich features that are acquired from other sub-aperture images. `$Conv, a, b, k$' represents 2D convolution with $a$ number of input channels, $b$ number of output channels, and $k$ is the kernel size.}
    \label{fig: model}
    \vspace{-15pt}
\end{figure}

\vspace{-5pt}
\section{Methodology}
\label{sec:method}

\vspace{-5pt}
\subsection{Motivation and Problem Formulation}
\label{subsec:problem}

Well-developed research has been achieved in the single-image super-resolution domain, leading to extraordinary performance in a single image. Images captured by the LF camera can be up-scaled by pre-trained SISR models by considering each sub-aperture image as a single image. However, it fails to exploit the angular and disparity information present in multiple SAIs. Our main objective is to propose an LF domain-specific module on top of the SISR model to achieve LFSR. Recently developed SISR models, $F_{SR}$ can be divided into two parts. One is feature extraction module $F_{feat}$, and another one is upscaling cum reconstruction module $F_{up}$. $F_{feat}$ extracts the salient features from a single image that is up-scaled by $F_{up}$. Our main objective is to introduce a module that modulates the extracted features from $F_{feat}$ by exploiting angular information across SAIs.

\vspace{-5pt}
\subsection{Light-field Sub-aperture Feature Adaptation Module}
\label{subsec:module}

Using our proposed adaptation module, the pre-trained SISR model adapts the spatial and angular information present in LF images which eventually improves the super-resolution performance further, as shown in Fig.~\ref{fig: algo}. The feature extraction and upscale module are from any pre-trained SISR model, and we place the adaptation module between them. Fig.~\ref{fig: model} shows the proposed light-field domain information adaptation module. The light-field camera captures multiple SA images of the same scene. In this work, the rich information present in those angular SA images is utilized to enhance each SA image's quality. $f_{i}$ is extracted feature set from a sub-aperture image $I_{i}$ using pre-trained feature extraction module $F_{feat}$ of a SISR model. $f_1, f_2, ...f_n$ are extracted features from $n$ different sub-aperture images. Each sub-aperture feature set is processed through its corresponding Sub-Aperture Shift (SAS) module. Each SAS module is expected to process the features in such a way that it can improve the performance on the sub-aperture image $I_{i}$ task in hand. The term shift in SAS is used as it is desired that the SAS module is expected to align the features from different sub-aperture images in the direction of the SA image feature set in hand to improve the performance. Fusion block $F_s$ fuses the processed SA features and feeds the fused features $f_i^{'}$ into the upscale module. During training, the weights of the pre-trained feature extraction and upscale modules are not updated during gradient back-propagation. We only update the weights of the adaptation module, as shown in Fig.~\ref{fig: algo}.

\textbf{SAS Module: } If we consider a light-field image of $a\times a$ angular resolution, there will be $n = a^2$ number of SA images. Therefore,  we have $n$ SAS modules for $n$ SA images. Each module takes its corresponding extracted SA features concatenated with a difference feature map. The difference feature map represents the difference between the SA feature set in hand and the SA feature set of that corresponding SAS module. The difference feature map helps to shift or modify a SA feature set in such a way that it will improve the performance of the SA feature set in hand. The difference map acts as a modulator that decides how much shift is required for a SA feature for pixel-wise mapping. The output of a SAS module can be mathematically represented as
\begin{equation}
    f_i^j = SAS_j([f_j, f_i-f_j]), j\in1, 2, ...,n
    \label{eq:sas}
\end{equation}
$f_i$ is the extracted features of $i^{th}$ angular SA image, which will be super-resolved, and $f_j$ is the extracted features of $j^{th}$ angular SA image. Both $f_j$ and $f_i-f_j$ are concatenated before feeding into $SAS_j$ module. This module is expected shift features $f_j$ and align with $f_i$ using the difference map $f_i-f_j$. All the SAS modules will align the features with $f_i$ and feed them into the fusion module.

\textbf{Fusion Module: } All the modulated SA features are fused together. It is mathematically expressed as
\begin{equation}
    f_i^{'} = f_i + F_s([ f_i^1, f_i^2, ..., f_i^n])
\end{equation}
$F_s$ is the fusion module and $f_i^{'}$ is the sub-aperture informative fused feature of $i^{th}$ SA feature $f_i$.


\vspace{-5pt}
\subsection{Architecture Details}
\label{subsec:archi}
Each SAS module consists of one convolution block and three consecutive residual blocks, as shown in Fig.~\ref{fig: model}. Fusion block contains two convolution layers. The first convolution layer blends all the SA features using $1\times1$ convolution, and the second convolution layer processes the fused features. We consider the popular RDN~\cite{RDN} and EDSR~\cite{EDSR} architecture of SISR for experimental purposes. In both cases, there are $64$ features in the feature set that are extracted from the feature extraction block. Therefore, we consider $C_i=64$ and $C_x=32$ in our experimental setup, as shown in Fig.~\ref{fig: model}.

\begin{table}
\caption{PSNR$/$SSIM values achieved by different methods for $2\times$ and $4\times$SR. Our results are shown in bold.}\label{tabComparison}
\vspace{-5pt}
\centering
\renewcommand\arraystretch{1.1}
\resizebox{0.47\textwidth}{!}{
\begin{tabular}{|l|c|lll|}
\hline
\multirow{2}*{Method}&\multirow{2}*{Scale}&  \multicolumn{3}{c|}{Dataset} \\
\cline{3-5}
  &  & EPFL \cite{EPFL} & INRIA \cite{INRIA} & STFgantry \cite{STFgantry} \\
\hline
Bicubic & $2\times$  & 29.50$/$0.935 &  31.10$/$0.956 & 30.82$/$0.947\\
VDSR \cite{VDSR}    & $2\times$ & 32.50$/$0.960 & 34.43$/$0.974 & 35.54$/$0.979\\
RCAN \cite{RCAN}    & $2\times$ & 33.16$/$0.964 & 35.01$/$0.977 & 36.33$/$0.983\\
resLF \cite{resLF}  & $2\times$ & 32.75$/$0.967 & 34.57$/$0.978 & 36.89$/$0.987\\
LFSSR \cite{LFSSR}  & $2\times$ & 33.69$/$0.975 & 35.27$/$0.983 & 38.07$/$0.990\\
LF-InterNet \cite{LF-InterNet}  & $2\times$ & 34.14$/$0.976 & 35.80$/$0.985 & 38.72$/$0.992\\
LF-DFnet \cite{wang2020light}  & $2\times$ & 34.44$/$0.977 &  36.36$/$0.984 & 39.61$/$0.994\\
MEG-Net \cite{zhang2021end} & $2\times$ & 34.31$/$0.977 &  36.10$/$0.985 & 38.77$/$0.992\\
DPT \cite{DPT} & $2\times$ & 34.49$/$0.976 &  36.41$/$0.984 & 39.43$/$0.993\\
LFSAFA-EDSR & $2\times$ &\textbf{35.08$/$0.973} &  \textbf{37.51$/$0.983} & \textbf{38.69$/$0.990}\\
LFSAFA-RDN  & $2\times$ &\textbf{35.19$/$0.974} &  \textbf{37.64$/$0.983} & \textbf{39.02$/$0.991}\\
\hline
Bicubic & $4\times$  & 25.14$/$0.831 &  26.82$/$0.886 & 25.93$/$0.843 \\
VDSR \cite{VDSR} & $4\times$ & 27.25$/$0.878 & 29.19$/$0.921 & 28.51$/$0.901 \\
RCAN \cite{RCAN} & $4\times$ & 27.88$/$0.886 & 29.76$/$0.927 & 28.90$/$0.911\\
resLF \cite{resLF}  & $4\times$ & 27.46$/$0.890 &  29.64$/$0.934 & 28.99$/$0.921\\
LFSSR \cite{LFSSR}  & $4\times$ & 28.27$/$0.908 &  30.31$/$0.945 & 30.15$/$0.939\\
LF-InterNet \cite{LF-InterNet} & $4\times$ & 28.67$/$0.914 &  30.64$/$0.949 & 30.53$/$0.943\\
LF-DFnet \cite{wang2020light} & $4\times$ &28.77$/$0.917 &  30.83$/$0.950 & 31.15$/$0.949\\
MEG-Net \cite{zhang2021end} & $4\times$ & 28.75$/$0.916 &  30.67$/$0.949 & 30.77$/$0.945\\
DPT \cite{DPT}   & $4\times$ & 28.94$/$0.917 &  30.96$/$0.950 & 31.15$/$0.949\\
LFSAFA-EDSR & $4\times$ &\textbf{29.47$/$0.909} &  \textbf{31.88$/$0.945} & \textbf{30.41$/$0.937}\\
LFSAFA-RDN  & $4\times$ &\textbf{29.62$/$0.911} &  \textbf{32.06$/$0.947} & \textbf{30.80$/$0.941}\\
\hline
\end{tabular}
}
\vspace{-15pt}
\end{table}

\vspace{-5pt}
\section{Experiments}
\label{sec:experiment}
\vspace{-5pt}
\subsection{Implementation Details}
\label{subsec:implement}

We used images from $5$ publicly available LF datasets, namely EPFL~\cite{EPFL}, HCInew~\cite{HCInew}, HCIold~\cite{HCIold}, INRIA~\cite{INRIA}, and STFgantry~\cite{STFgantry}. We follow the same train-test split as given by~\cite{wang2020light}. There are a total of $144$ training images in the dataset and consider standard $5\times 5$ angular resolution for benchmark analysis. For testing, we use real-world images from EPFL~\cite{EPFL}, INRIA~\cite{INRIA}, and STFgantry~\cite{STFgantry} datasets which consists of $10$, $5$, and $2$ test images, respectively. EDSR~\cite{EDSR} and RDN~\cite{RDN} are the base SISR models, where we insert our proposed LFSAFA module to adapt these models for light-field imaging. Bicubic downsampling generates low-resolution (LR) images from its high-resolution (HR) counterpart. We extract random $32\times32$ crop from LR images during training and augment the patch using random $90^\circ$ rotation with a random horizontal and vertical flip. We train the LFSAFA module for $250$ epochs, and each epoch consists of $\sim 1000$ batch updates with a batch size of $4$. Adam optimizer with a learning rate $10^{-4}$ is used for updating the weights, and the learning rate is reduced by a factor of $0.5$ after every $50$ epochs. The mean absolute difference between output reconstruction and HR image is employed as a loss function. Peak signal-to-noise ratio (PSNR) and Structural Similarity Index (SSIM) are calculated on all the sub-aperture views for comparative performance analysis. Larger values on those metrics imply better reconstruction performance. Following the trend in the LFSR domain, PSNR and SSIM are calculated on the luminance channel Y of an image in YCbCr space. The code is available at \href{https://aupendu.github.io/LFSAFA-SR}{https://aupendu.github.io/LFSAFA-SR}.
\vspace{-5pt}
\subsection{Performance Analysis}
\label{subsec:perform}

\begin{figure*}%
    \centering
    \subfloat{{\includegraphics[width=2cm]{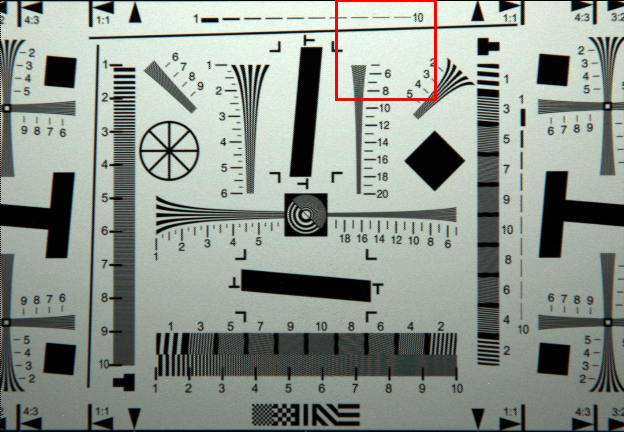} }}%
    \subfloat{{\includegraphics[width=2cm]{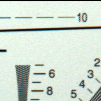} }}%
    \subfloat{{\includegraphics[width=2cm]{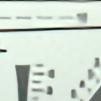} }}%
    \subfloat{{\includegraphics[width=2cm]{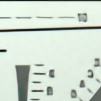} }}%
    \subfloat{{\includegraphics[width=2cm]{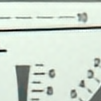} }}%
    \subfloat{{\includegraphics[width=2cm]{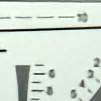} }}%
    \subfloat{{\includegraphics[width=2cm]{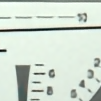} }}%
    \subfloat{{\includegraphics[width=2cm]{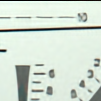} }}%
    \vspace{-10pt}
    
    \subfloat{{\includegraphics[width=2cm]{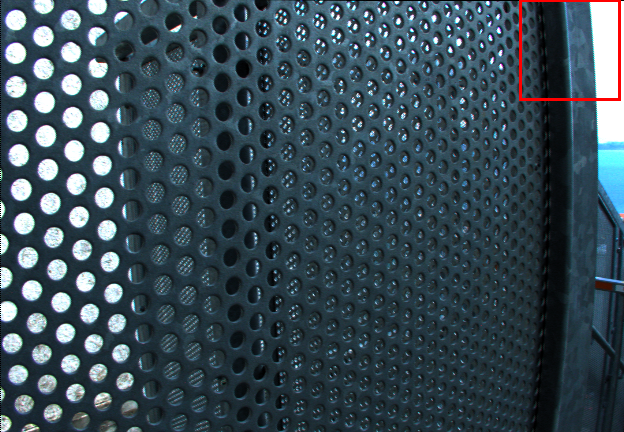} }}%
    \subfloat{{\includegraphics[width=2cm]{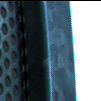} }}%
    \subfloat{{\includegraphics[width=2cm]{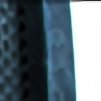} }}%
    \subfloat{{\includegraphics[width=2cm]{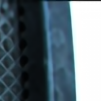} }}%
    \subfloat{{\includegraphics[width=2cm]{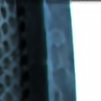} }}%
    \subfloat{{\includegraphics[width=2cm]{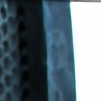} }}%
    \subfloat{{\includegraphics[width=2cm]{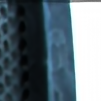} }}%
    \subfloat{{\includegraphics[width=2cm]{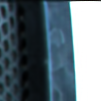} }}%
    \vspace{-10pt}
    \setcounter{subfigure}{0}
    
    \subfloat[\centering Whole Image]{{\includegraphics[width=2cm]{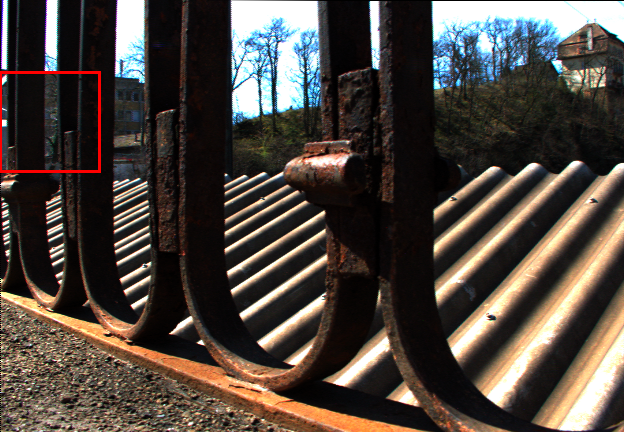} }}%
    \subfloat[\centering Cropped HR]{{\includegraphics[width=2cm]{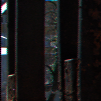} }}%
    \subfloat[\centering VDSR]{{\includegraphics[width=2cm]{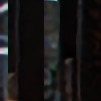} }}%
    \subfloat[\centering RCAN]{{\includegraphics[width=2cm]{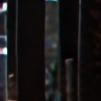} }}%
    \subfloat[\centering LF-InterNet]{{\includegraphics[width=2cm]{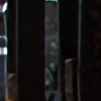} }}%
    \subfloat[\centering MEG-Net]{{\includegraphics[width=2cm]{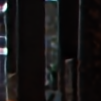} }}%
    \subfloat[\centering DPT]{{\includegraphics[width=2cm]{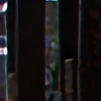} }}%
    \subfloat[\centering Ours]{{\includegraphics[width=2cm]{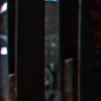} }}%
    \vspace{-5pt}
    \caption{Qualitative comparison of our proposed LFSAFA-RDN with the existing LFSR algorithms for $4\times$ SR.}%
    \label{fig:comp_pic}%
    \vspace{-15pt}
\end{figure*}

We compare the performance of our proposed approach with state-of-the-art single image super-resolution and light field super-resolution models. VDSR~\cite{VDSR}, RCAN~\cite{RCAN} are the SISR models, and the rest of the methods in Table~\ref{tabComparison} are the LFSR models. All the SISR models are trained on light field training images for a fair comparison. We can observe from Table~\ref{tabComparison} that both the SISR models with our proposed modification outperform all the existing techniques in terms of PSNR on EPFL and INRIA datasets. The proposed method cannot outperform other methods in the SSIM metric. However, the SSIM values are very close to the best. Fig.~\ref{fig:comp_pic} shows the qualitative comparison of our proposed algorithm with existing LFSR approaches. We observe that our proposed algorithm achieves a more satisfactory reconstruction of numbers in the first-row image, excellently reconstructs the round holes in the second-row image, and adequately preserves the line structure, which is on the left side of the third-row image.

\begin{table}[]
\centering
\caption{Model ablation studies of our proposed LFSAFA module and the effect of angular resolution on the reconstruction performance. All the experiments are performed on the LFSAFA-RDN variant for $2\times$ SR.}
\vspace{-5pt}
\label{tab:ablation}
\resizebox{0.47\textwidth}{!}{%
\begin{tabular}{|c|c|c|c|c|c|}
\hline
 &
  \begin{tabular}[c]{@{}c@{}}Residual\\ Connection\end{tabular} &
  \begin{tabular}[c]{@{}c@{}}Difference \\ Feature\end{tabular} &
  \begin{tabular}[c]{@{}c@{}}Adaptation\\ Module\end{tabular} &
  \begin{tabular}[c]{@{}c@{}}Angular\\ Resolution\end{tabular} &
  PSNR/ SSIM \\ \hline
Ablation-1 & \ding{55}  & \ding{55}  & \ding{55}  & $3\times3$ & \multicolumn{1}{l|}{34.10/ 0.9662} \\ \hline
Ablation-2 & \ding{51} & \ding{55}  & \ding{51} & $3\times3$ & 34.17/ 0.9664                      \\ \hline
Ablation-3 & \ding{55}  & \ding{51} & \ding{51} & $3\times3$ & 34.81/ 0.9711                      \\ \hline
Ablation-4 & \ding{51} & \ding{51} & \ding{51} & $3\times3$ & 34.86/ 0.9715                      \\ \hline
Ablation-5 & \ding{51} & \ding{51} & \ding{51} & $5\times5$ & 35.19/ 0.9737                      \\ \hline
\end{tabular}%
}
\vspace{-15pt}
\end{table}
\vspace{-5pt}
\subsection{Ablation Studies}
\label{subsec:ablation}

Table~\ref{tab:ablation} shows the model ablation studies of our proposed LFSAFA module. Along with the model ablation, a study on the effect of the angular resolution is given in that table. All the components are the same for both Ablation-4 and Ablation-5 except the angular resolution of the input LF image. We can observe from the table that the network's performance in terms of PSNR and SSIM increases as we increase the angular resolution. The proposed module LFSAFA gets more angular information as we increase the angular resolution. Therefore, it leads to better performance. This phenomenon also supports our claim that our module has the potential to explore sub-aperture angular information. Other ablations show the effectiveness of three different parts of the LFSAFA module. The first one is the residual connection between the input and output, the second one is the inclusion of difference features into each SAS module, and the third one is the contribution of the whole proposed adaptation module, LFSAFA. The only difference between Ablation-3 and Ablation-4 is the residual connection, and it shows that performance improves slightly with that connection, and we also observe that the network converges faster. Ablation-1 is basically the SISR model without the LFSAFA module. The performance does not improve much even if we add the proposed adaptation module without the difference feature, as shown in Ablation-2. Therefore, the difference feature plays a key role, and we can observe that by comparing the Ablation-2 and Ablation-4. There is a significant jump in performance metrics. Therefore, we can say that the difference feature plays a crucial role in utilizing the sup-aperture information in a controlled manner. Table~\ref{tab:ab2} shows the summarized main contribution of this paper. LFSAFA module helps both the SISR models to adopt the LF domain-specific extra angular information. We can observe a significant $\sim 1 dB$ improvement in PSNR metric across all the datasets and models for $2\times$ SR, and $\sim 0.6-1 dB$ improvement for $4\times$ SR.

\begin{table}[]
\centering
\caption{Comparative analysis of our proposed LFSAFA module-based LFSR models with their SISR counterparts.}
\vspace{-5pt}
\label{tab:ab2}
\resizebox{0.47\textwidth}{!}{%
\begin{tabular}{|c|l|cc|cc|}
\hline
Scale               & \multicolumn{1}{c|}{Dataset} & RDN            & LFSAFA-RDN     & EDSR           & LFSAFA-EDSR    \\ \hline
\multirow{3}{*}{$\times 2$} & EPFL                         & 34.14$/$0.966 & 35.19$/$0.974 & 34.06$/$0.966 & 35.08$/$0.973 \\ \cline{2-2}
                    & INRIA                        & 36.42$/$0.978 & 37.64$/$0.983 & 36.28$/$0.978 & 37.51$/$0.983 \\ \cline{2-2}
                    & STFgantry                    & 37.91$/$0.987 & 39.02$/$0.991 & 37.48$/$0.986 & 38.69$/$0.990 \\ \hline
\multirow{3}{*}{$\times 4$} & EPFL                         & 28.84$/$0.898 & 29.62$/$0.911 & 28.73$/$0.895 & 29.47$/$0.909 \\ \cline{2-2}
                    & INRIA                        & 31.06$/$0.935 & 32.06$/$0.947 & 30.92$/$0.933 & 31.88$/$0.945 \\ \cline{2-2}
                    & STFgantry                    & 30.18$/$0.932 & 30.80$/$0.941 & 29.75$/$0.926 & 30.41$/$0.937 \\ \hline
\end{tabular}%
}
\vspace{-15pt}
\end{table}

\vspace{-5pt}
\section{Conclusion}
\label{sec:conclude}
In this work, we propose a module that will turn a SISR model into an LFSR model. The proposed module can be used in all the recently developed SISR models without architectural modifications. This paper presents a new research direction in the LFSR domain, which will drive the community to develop a better sub-aperture feature adaptation module. In the future, a more powerful sub-aperture feature adaptation module can improve the performance further.

\bibliographystyle{IEEEbib}
\bibliography{refs.bib}

\end{document}